\documentclass{article} 
\usepackage{iclr2027_conference,times}

\usepackage{amsmath,amsfonts,bm}

\def\eqref#1{equation~\ref{#1}}

\def\1{\bm{1}}

\DeclareMathAlphabet{\mathsfit}{\encodingdefault}{\sfdefault}{m}{sl}
\SetMathAlphabet{\mathsfit}{bold}{\encodingdefault}{\sfdefault}{bx}{n}

\usepackage{amsmath}
\usepackage{amssymb}
\usepackage{booktabs}
\usepackage{graphicx}
\usepackage{xcolor}
\usepackage{array}
\usepackage{longtable}
\usepackage{hyperref}
\usepackage{url}
\usepackage{xspace}
\definecolor{contributionbg}{RGB}{239,246,248}
\definecolor{contributionedge}{RGB}{16,82,99}

\title{Energy-guided Recursive Model}

\author{
  \makebox[0pt][c]{Yifei Zhao$^{1,}$\thanks{Correspondence authors: \texttt{202521210324@std.uestc.edu.cn}} \quad Ying Tang$^{1,2}$\thanks{Correspondence authors: \texttt{jamestang23@gmail.com}}} \\
  \\
  \makebox[0pt][c]{$^1$Institute of Fundamental and Frontier Sciences,} \\
  \makebox[0pt][c]{University of Electronic Science and Technology of China, Chengdu 611731, China} \\
  \makebox[0pt][c]{$^2$School of Physics, University of Electronic Science and Technology of China, Chengdu 611731, China}
}

\iclrfinalcopy 
\renewcommand{\lhead}[2][]{}
\renewcommand{\headrulewidth}{0pt}
\begin{document}
\maketitle

\begin{abstract}
Recursive models show promise on reasoning and language tasks, yet their test-time scaling lacks a principled criterion for selecting trajectories or determining recurrent depth. We introduce \textbf{Energy-guided Recursive Model (ERM)}, which uses Hopfield-type memories of valid local and global structures to assign intrinsic energies to candidate trajectories. These energies guide candidate selection and suggest an effective range of recurrent depths, implying that deeper recurrence does not necessarily improve reasoning accuracy. They also enable sampling methods such as parallel tempering to improve exploration. For reasoning tasks, ERM achieves optimal solutions on Sudoku ($98.97\%$), Pencil Puzzle Bench (PPBench, $88.04\%$) and Maze ($99.30\%$), reaching the best accuracy in recursive modeling. On language modeling, ERM reduces RedPajama-V2 perplexity by $1.74\%$ with marginal inference overhead. The results support energy guidance as a practical framework for improving test-time scaling in recursive models.
\end{abstract}

\begin{center}
\includegraphics[width=0.90\linewidth]{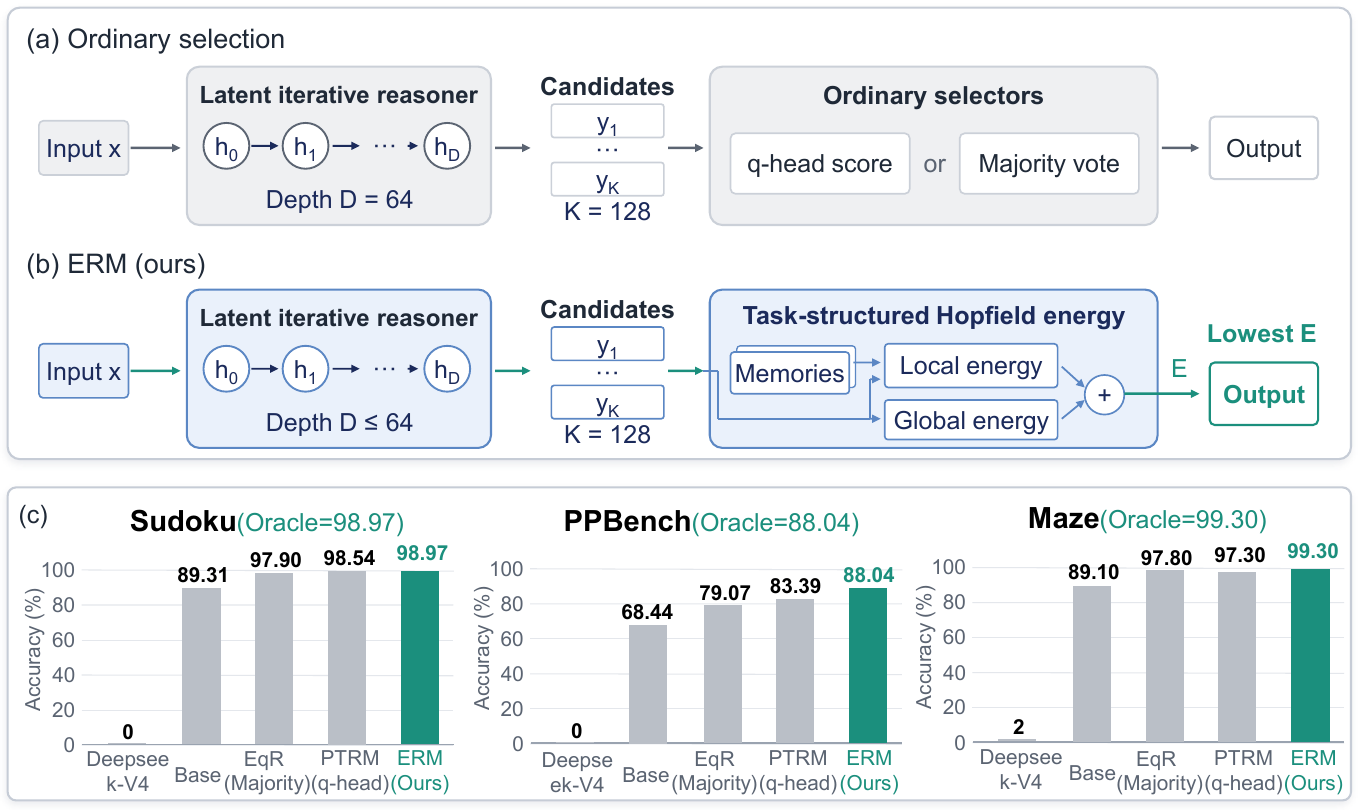}
\end{center}
{
\refstepcounter{figure}\label{fig:energy_framework}
\noindent Figure~\thefigure: \textbf{Energy-guided recursive model improves on reasoning tasks.} (a) A reasoner generates $K$ candidates at depth $D$; q-head scores or majority voting select the output. (b) ERM combines local Hopfield energies with a global constraint penalty and selects the lowest-energy candidate, without additional training. (c) Accuracy on three representative reasoning tasks. The ERM bars match their respective candidate-pool oracles.}

\section{Introduction}

Test-time computation improves recursive models only when additional work produces better predictions, in both reasoning and language tasks. Autoregressive language models use longer chains of thought \citep{wei2022chain}, multiple rationales \citep{wang2023selfconsistency}, search \citep{yao2023tree}, or learned verifiers \citep{cobbe2021verifiers,lightman2024verify} to exploit additional inference budget \citep{snell2025scaling}. Recursive models generate candidate trajectories, but more trajectories or greater depth do not ensure better predictions. The shared problem is how to select useful candidates and decide when further recurrence helps.

Loop models reuse a shared operator within one forward computation, as in Neural GPUs~\citep{kaiser2016neuralgpus}, adaptive computation time (ACT)~\citep{graves2016adaptive}, Universal Transformers~\citep{dehghani2019universal}, and implicit or equilibrium networks~\citep{geng2021trainingimplicit,bai2019deep}. Recent transformers use recurrence for iterative algorithms, latent reasoning, and adaptive depth \citep{giannou2023looped,yang2024looped,saunshi2025latent,geiping2025latent,schone2025implicit,bae2025mixture,zhu2025ouro}. HRM, TRM, EqR, and PTRM apply compact recurrent networks to structured reasoning \citep{wang2025hrm,jolicoeurmartineau2025trm,huang2026eqr,sghaier2026ptrm}. Their trajectories can encounter spurious fixed points \citep{ren2026guessing}, making candidate selection consequential. Yet learned halting scores, confidence, and voting need not track prediction quality across recurrent steps.

Energy-based models assign scalar compatibility scores to configurations. Hopfield networks, Boltzmann machines, and energy-based learning define inference through low-energy states \citep{hopfield1982neural,ackley1985learning,lecun2006tutorial}; dense associative memories and modern Hopfield networks connect retrieval to attention \citep{krotov2016dense,ramsauer2020hopfield}. Energy Transformer further designs representation updates that descend an attention-based energy \citep{hoover2023energy}. We use Hopfield-type memories to define Hopfield energies over candidate predictions. Known rules supply these memories for reasoning, while observed continuations supply them for language modeling.

We introduce the \textbf{Energy-guided Recursive Model (ERM)}, a common framework for energy guidance in recursive reasoning and language models. Hopfield energies guide selection, sampling, and recurrent depth without further neural training (Figure~\ref{fig:energy_framework}). Our main contributions are:

\begingroup
\setlength{\fboxsep}{0pt}
\noindent\colorbox{contributionbg}{%
  {\color{contributionedge}\vrule width 2pt}%
  \hspace{8pt}%
  \begin{minipage}{\dimexpr\linewidth-18pt\relax}
  \vspace{6pt}
  \begin{list}{\textbullet}{\setlength{\leftmargin}{12pt}\setlength{\labelsep}{5pt}\setlength{\topsep}{0pt}\setlength{\parsep}{0pt}\setlength{\itemsep}{4pt}}
  \item \textbf{A general framework for energy guidance.} ERM uses Hopfield-type memories to define Hopfield energies for recursive reasoning and language models, supporting selection, sampling, and depth control.
  \item \textbf{Oracle accuracy on reasoning tasks.} With $D\leq64$ and $K=128$, ERM matches the evaluated candidate-pool oracles on Sudoku, PPBench, and Maze, and supports early stopping.
  \item \textbf{Improved language modeling.} The same framework uses Hopfield-type memories of observed continuations to reduce RedPajama-V2 perplexity by $1.74\%$ without further training.
  \end{list}
  \vspace{6pt}
  \end{minipage}%
  \hspace{8pt}%
}\par
\endgroup

\section{Related Work}

Latent iterative reasoning models reuse compact networks to solve structured tasks. HRM and TRM demonstrate this approach on Sudoku, Maze, and ARC-style problems \citep{wang2025hrm,jolicoeurmartineau2025trm}, while EqR and PTRM study attractor learning and stochastic exploration \citep{huang2026eqr,sghaier2026ptrm}. Solve the Loop and fixed-point reasoners relate recurrence to convergence and adaptive stopping \citep{feinashley2026solveloop,movahedi2026fprm}. Lai et al.\ link slow reasoning to transient chaos near saddle points representing nearly correct solutions, with increasingly fractal basins on harder tasks \citep{lai2026fractal}. These studies explain the dynamics and limits of additional iterations. Adaptive computation time (ACT) learns a halting policy~\citep{graves2016adaptive}; ERM instead tests whether an explicit task energy can both select candidates and supply a stopping signal.

Recent flow and diffusion models improve recurrent generation through denoising objectives. Flow Reasoning Models use self-conditioning and Fixed-Point Forcing to train on states produced by their own inference dynamics \citep{helbling2026flow}. Drozdova et al.\ retain a hidden state across denoising steps and remove timestep conditioning, allowing a single trajectory to refine solutions without candidate selection \citep{drozdova2026diffusion}. Looped flows train recurrent states with local denoising objectives and perform inference by probability-flow integration \citep{suleymanzade2026looped}. These methods improve the generator through training; ERM uses Hopfield energies to improve predictions from a fixed generator.

Energy-based models describe compatibility through a scalar energy. Classical Hopfield networks and Boltzmann machines associate low energy with preferred configurations \citep{hopfield1982neural,ackley1985learning,lecun2006tutorial}, while modern Hopfield networks connect memory retrieval to attention \citep{ramsauer2020hopfield}. Energy Transformer and IREM use energy minimization within representation updates or reasoning \citep{hoover2023energy,du2022irem}. Energy-Based Transformers (EBTs) learn an input--prediction energy and refine predictions through its gradients \citep{gladstone2026ebt}. Our language experiment keeps the EBT fixed and compares its learned scalar energy with Hopfield energies on the same candidates.

Verifier and reranking methods differ in the evidence used to judge predictions. Confidence and q-head selectors use model scores, while majority voting uses agreement among candidates \citep{sghaier2026ptrm}. Learned verifiers can use additional outcome or process supervision \citep{cobbe2021verifiers,lightman2024verify}, and reinforcement learning can improve the generator itself through reward-based parameter updates \citep{guo2025deepseek}. ERM requires neither a separately trained verifier nor post-training updates: its scores come from task memories or observed language continuations. This distinction motivates evaluating both the gain and the additional inference cost.

\section{Method}

\subsection{Problem Setup}

We study energy guidance for recursive models on reasoning and language tasks. Let $x$ denote an input puzzle or text prefix, and let a recursive model with depth $D$ produce $K$ candidate outputs $\mathcal{Y}(x)=\{y_1,\ldots,y_K\}$ through independent rollouts, perturbations, or parallel-tempering chains.  Each rollout provides a decoded prediction $y_k$ over $N$ positions and its predictive log-probabilities $L_k$. We write $E(y_k;x)$ as shorthand for a score that may also use $L_k$: Sudoku memories use these probabilities, while the other puzzle factors score decoded structures. Lower energy indicates greater task compatibility. For structured reasoning, the selected answer is
\begin{equation}
\hat{y}
= y_{\hat{k}},
\qquad
\hat{k}
= \arg\min_{k\in\{1,\ldots,K\}} E(y_k;x).
\label{eq:selection}
\end{equation}
This equation also defines the common baselines as energies: confidence uses negative mean token confidence, q-head uses the negative learned halting logit, and majority voting uses a negative agreement score.  ERM constructs $E$ from Hopfield-type memories and any global constraints. For language modeling, the same construction defines energies over tokens and combines candidate distributions, as described below.  Appendix~\ref{app:erm_pseudocode} gives the independent-rollout selection algorithm; PT additionally changes candidate generation.

\subsection{Energy-Guided Recursive Model}

ERM is derived from the Modern Hopfield view of energy-based memory retrieval.  In a Modern Hopfield network, a query state $q$ is compared with a memory set $\mathcal{M}$, and retrieval lowers an energy when $q$ is similar to at least one stored memory \citep{ramsauer2020hopfield}.  A compact form of this retrieval energy is
\begin{equation}
E_{\mathrm{MHN}}(q;\mathcal{M})
=
\frac{1}{2}\lVert q\rVert^2
-
\tau
\log
\sum_{m\in\mathcal{M}}
\exp\!\left(
\frac{\mathrm{sim}(q,m)}{\tau}
\right),
\label{eq:mhn_energy}
\end{equation}
where $\mathrm{sim}(q,m)$ is the similarity between the query and memory, $\tau$ is the retrieval temperature, and the squared-norm term regularizes the continuous query state.  ERM uses this log-sum-exp construction to define Hopfield energies over candidate predictions: each factor $j$ has an input-conditioned memory set $\mathcal{M}_j(x)$, the similarity is replaced by the negative distance $-\beta d_j(y_k,m)$, and constraints that are global rather than locally enumerable are written as $G(y_k,x)$:
\begin{equation}
E_{\mathrm{ERM}}(y_k;x)
=
\mu G(y_k,x)
-
\sum_{j=1}^{J}
\tau
\log
\sum_{m\in\mathcal{M}_j(x)}
\exp\!\left(
-
\frac{\beta d_j(y_k,m)}{\tau}
\right)
\label{eq:erm_energy}
\end{equation}
Here $d_j(y_k,m)$ measures mismatch to memory $m$, $\beta$ is the distance scale, and $\mu$ weights the global term. As $\tau$ approaches zero, each retrieval term approaches the minimum scaled distance to its local memory set. The factor sum therefore scores compatibility with enumerable local structures. The global penalty $G(y_k,x)$ accounts for additional constraints such as connectivity; it is omitted when no such constraints are specified. Reasoning uses memories of valid patterns, while language uses distributions over observed continuations.

\subsection{Parallel Tempering over Candidate Energies}

As an optional sampling extension, parallel tempering (PT) uses the same energy to define
\begin{equation}
\pi_T(y\mid x)
\propto
\exp\!\left(-\frac{E(y;x)}{T}\right),
\label{eq:tempered}
\end{equation}
where $T>0$ controls exploration. Replicas at $T_1<\cdots<T_R$ exchange states between neighboring temperatures, with acceptance probability
\begin{equation}
a
=
\min\left\{
1,
\exp\left[
    \left(\frac{1}{T_\rho}-\frac{1}{T_\eta}\right)(E_\rho-E_\eta)
\right]
\right\}.
\label{eq:pt_swap}
\end{equation}
Here $E_\rho$ and $E_\eta$ are the energies at temperatures $T_\rho$ and $T_\eta$. We evaluate PT through selected and oracle accuracy, without assuming that the recurrent dynamics exactly sample $\pi_T$.

\subsection{Reasoning Tasks}

The three reasoning tasks use Equation~\ref{eq:erm_energy}, with task-specific memories, distances, and global constraints. In the shared-pool experiments, each selector uses the same generated candidates; in PT, the energy also guides replica exchanges during generation.

Sudoku instantiates $E_{\mathrm{ERM}\mid\mathrm{Sudoku}}$ with row, column, and box memories.  Each factor $j$ is one row, column, or $3\times3$ box, and each memory $m\in\mathcal{M}_j(x)$ assigns one valid digit to each of the nine cells in that factor.  Let $L_k[i,d]$ be rollout $k$'s predictive log-probability for digit $d$ at cell $i$, and let $m_i$ be the digit in memory $m$.  The Sudoku energy is the direct specialization
\begin{equation}
E_{\mathrm{ERM}\mid\mathrm{Sudoku}}(y_k,L_k;x)
=
-\sum_j
\tau\log\sum_{m\in\mathcal{M}_j(x)}
\exp\!\left(
\frac{1}{\tau}\sum_{i\in j} L_k[i,m_i]
\right),
\label{eq:sudoku_energy}
\end{equation}
The sum over $j$ covers rows, columns, and boxes. The memory distance is the negative log-probability assigned to a valid unit pattern, so this score uses soft predictions rather than only the decoded board. A log-permanent dynamic program evaluates the inner sum exactly; clue cells are clamped to their input digits.

PPBench instantiates $E_{\mathrm{ERM}\mid\mathrm{PPBench}}$ with puzzle-type-specific local memories plus global distance potentials.  For enumerable rules, each factor $j$ stores legal binary or multi-class local patterns in $\mathcal{M}_j(x)$, and $d_j(y_k,m)$ is a normalized Hamming-style distance between the candidate pattern and a legal memory.  Lightup uses numbered-clue exact-count memories and line-of-sight at-most-one-bulb memories; Tapa uses cyclic run-pattern memories around clues; Heyawake uses room-count and adjacency memories; Nurikabe uses no-$2\times2$-sea memories and island/sea potentials; and the Sudoku subset reuses the Sudoku permutation memories.  Connectivity and coverage rules are not faithfully enumerable as small local memories, so they enter through $G(y_k,x)$:
\begin{equation}
E_{\mathrm{ERM}\mid\mathrm{PPBench}}(y_k;x)
=
64\,G(y_k,x)
-
\sum_j \tau\log\sum_{m\in\mathcal{M}_j(x)}
\exp\!\left(-\frac{128\,d_j(y_k,m)}{\tau}\right).
\label{eq:ppbench_energy}
\end{equation}
Here $\beta=128$ and $\mu=64$; no separate confidence score is added.

Maze instantiates $E_{\mathrm{ERM}\mid\mathrm{Maze}}$ with local path memories and a global rule penalty. The local factors use the same $\mathcal{M}_j(x)$ notation as the other tasks: terminal factors store patterns in which the start or goal has path degree one, ordinary open-cell factors store either off-path patterns or on-path degree-two patterns, and wall factors enforce consistency with the input maze. Disconnected cycles and fragments can satisfy local degree constraints, so $G(y_k,x)$ additionally penalizes violations of the global path rules. The resulting energy is
\begin{equation}
E_{\mathrm{ERM}\mid\mathrm{Maze}}(y_k;x)
=
G(y_k,x)
-
0.25\sum_j \tau\log\sum_{m\in\mathcal{M}_j(x)}
\exp\!\left(-\frac{d_j(y_k,m)}{\tau}\right).
\label{eq:maze_energy}
\end{equation}
The global rule penalty $G(y_k,x)$ checks format consistency with the input maze, start-goal connectivity, path degrees, cycles, and path length. The local retrieval weight is $0.25$, with unit distance and global-penalty scales.

\subsection{Language Modeling and Task Adaptation}

The same framework applies to both reasoning and language by specifying the prediction variables, Hopfield-type memories, and any global constraints. Reasoning uses valid local patterns and global rules. Language has many acceptable continuations, so its Hopfield-type memories store token distributions rather than discrete valid assignments. Hopfield energies then define a normalized prediction distribution.

\paragraph{Language prediction.}
We apply the same factor construction in Equation~\ref{eq:erm_energy} to token distributions. At position $t$, let $p_k(v)$ be the conditional probability of token $v$ from trajectory $k$, and let $q_t(v)$ be a continuation distribution retrieved using the observed prefix $x_{<t}$. Define
\begin{equation}
p_0(v)=\frac{1}{K}\sum_{k=1}^K p_k(v),\qquad
r(v)=(1-g_t)p_0(v)+g_t q_t(v),
\label{eq:language_mixture}
\end{equation}
where $g_t\in[0,1]$ controls the memory contribution. The two factors use $\mathcal M_1=\{p_1,\ldots,p_K\}$ with uniform weights $1/K$, and $\mathcal M_2=\{p_0,q_t\}$ with weights $(1-g_t,g_t)$. For a memory distribution $p_m$, the mismatch is $d_j(v,m)=-\log p_m(v)$. We average the two weighted retrieval energies at unit temperature:
\begin{equation}
E_{\mathrm{LM}}(v)=-\frac12\sum_{j=1}^{2}\log\!\sum_{m\in\mathcal M_j}
w_{j,m}\exp[-d_j(v,m)]+\log Z_t,
\label{eq:language_energy}
\end{equation}
where $Z_t=\sum_{u\in\mathcal V}\sqrt{p_0(u)r(u)}$ normalizes $\exp[-E_{\mathrm{LM}}(v)]$ over the vocabulary. The resulting token distribution is the normalized geometric mean of $p_0$ and $r$: the first factor retains model evidence, and the second combines it with retrieved continuations. At $g_t=0$, both factors reduce to $p_0$. The language formulation uses weighted Hopfield-type memories at $\tau=\beta=1$, averages two factors, and sets $\mu=0$.

\section{Experiments}

\subsection{Experimental Setup}

We evaluate ERM on reasoning and language tasks with fixed pretrained models. Reasoning experiments use EqR with $D\leq64$ and $K=128$; the shared-pool comparison fixes $D=64$, while depth and stopping studies use smaller budgets. Language experiments use an EBT with $K=8$ (Section~\ref{sec:language_results}). The reasoning suite contains Sudoku-Lite, the five-type PPBench validation split derived from Pencil Puzzle Bench \citep{waugh2026ppbench}, and the official EqR Maze-Unique setting.  The Maze task uses the same dataset used by EqR: a uniquely solvable Maze-hard-1k variant synthesized from perfect $30\times30$ mazes by sampling start-goal pairs, filtering path lengths, and de-duplicating train/test splits \citep{huang2026eqr}.  Reasoning comparisons report exact accuracy, requiring the full answer to match the target; appendix diagnostics also report per-position token accuracy.  Appendix Table~\ref{tab:param_compute} summarizes the parameter and compute accounting: ERM adds no learned neural parameters and only adds task-energy scoring on top of the shared rollout budget.  Language modeling uses token-weighted perplexity to assess the full prediction distribution.

\begin{figure}[t]
  \centering
  \includegraphics[width=\linewidth]{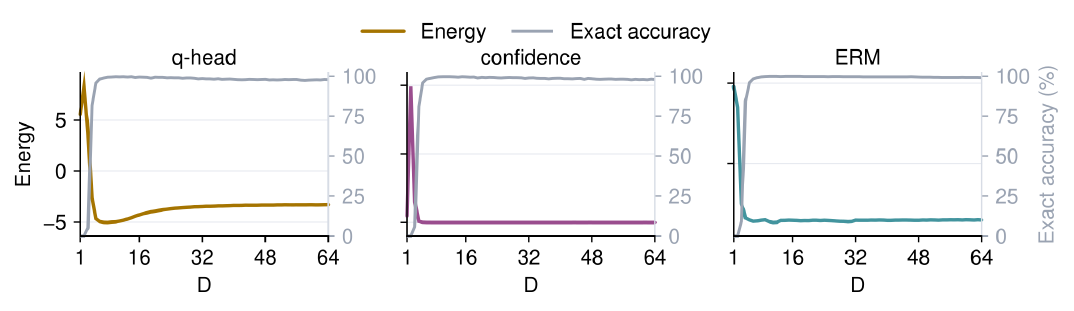}
  \caption{\textbf{Maze selector dynamics over recurrent depth.}
  The solid curve in each panel shows the mean selected energy, and the dashed gray curve uses the right axis to show exact accuracy as recurrent depth increases from $D=1$ to $D=64$ with $K=128$ Maze rollouts.  Confidence uses negative mean valid-token maximum log-probability, while q-head and confidence quickly become saturated selector scores.  ERM uses the Maze instantiation of Equation~\ref{eq:erm_energy}; its energy decreases smoothly over most of the evaluated depths.}
  \label{fig:maze_depth_curve}
\end{figure}

For reasoning, we compare ERM with four non-oracle selectors.  The one-rollout baseline uses a single EqR prediction.  Whole-sequence majority chooses the most frequent complete answer among the $K$ candidates when applicable, while cell-majority voting is additionally reported for PPBench and Maze diagnostics.  The q-head selector chooses the candidate with the largest learned EqR halting score, and confidence chooses the candidate with the largest mean valid-token confidence.  The candidate oracle uses the target to report whether any of the $K$ candidates is exactly correct, defining the selection upper bound for that pool.  Because Maze exposes a particularly sharp distinction between local confidence and global path compatibility, Figure~\ref{fig:maze_depth_curve} also shows how the selector energy and exact accuracy evolve over recurrent depth.

Section~\ref{sec:energy_stopping} tests whether the energy dynamics in Figure~\ref{fig:maze_depth_curve} can guide early stopping.

\begin{table}[t]
\caption{\textbf{Shared-candidate selection at $D=64$, $K=128$.}
All methods select from the same candidate pool for each task.  Base denotes the one-rollout baseline, Conf. denotes confidence selection, the oracle is the best possible exact accuracy inside that pool, and the ERM gap is oracle minus ERM.}
\label{tab:shared}
\centering
\small
\begin{tabular}{lrrrrrrr}
\toprule
Task & Base & Majority & q-head & Conf. & ERM & Oracle & Gap \\
\midrule
Sudoku & 89.31 & 97.90 & 98.54 & 98.58 & 98.83 & 98.83 & 0.00 \\
PPBench & 68.44 & 79.07 & 83.39 & 80.73 & 88.04 & 88.04 & 0.00 \\
Maze & 89.10 & 97.80 & 97.30 & 97.80 & 99.30 & 99.30 & 0.00 \\
\bottomrule
\end{tabular}
\end{table}

\subsection{Oracle Accuracy on Reasoning Tasks}

On Sudoku, ERM selects all $2024/2048$ correct answers available in the shared pool ($98.83\%$), five more than confidence selection. The remaining errors arise from candidate generation; PT adds three further solves (Table~\ref{tab:pt}).

PPBench shows the largest selection gain: ERM solves $265/301$ puzzles, matching the oracle and exceeding q-head by 14 solves. Its local memories and global penalties distinguish valid candidates across heterogeneous puzzle rules.

On Maze, ERM reaches the $99.30\%$ oracle, compared with $97.80\%$ for majority and confidence and $97.30\%$ for q-head. Thus ERM recovers every correct answer available in this pool. At $D=16$, both ERM and the oracle reach $99.90\%$ (Table~\ref{tab:maze_stopping}); deeper recurrence therefore does not guarantee better candidate coverage.

\begin{figure}[t]
  \centering
  \includegraphics[width=\linewidth]{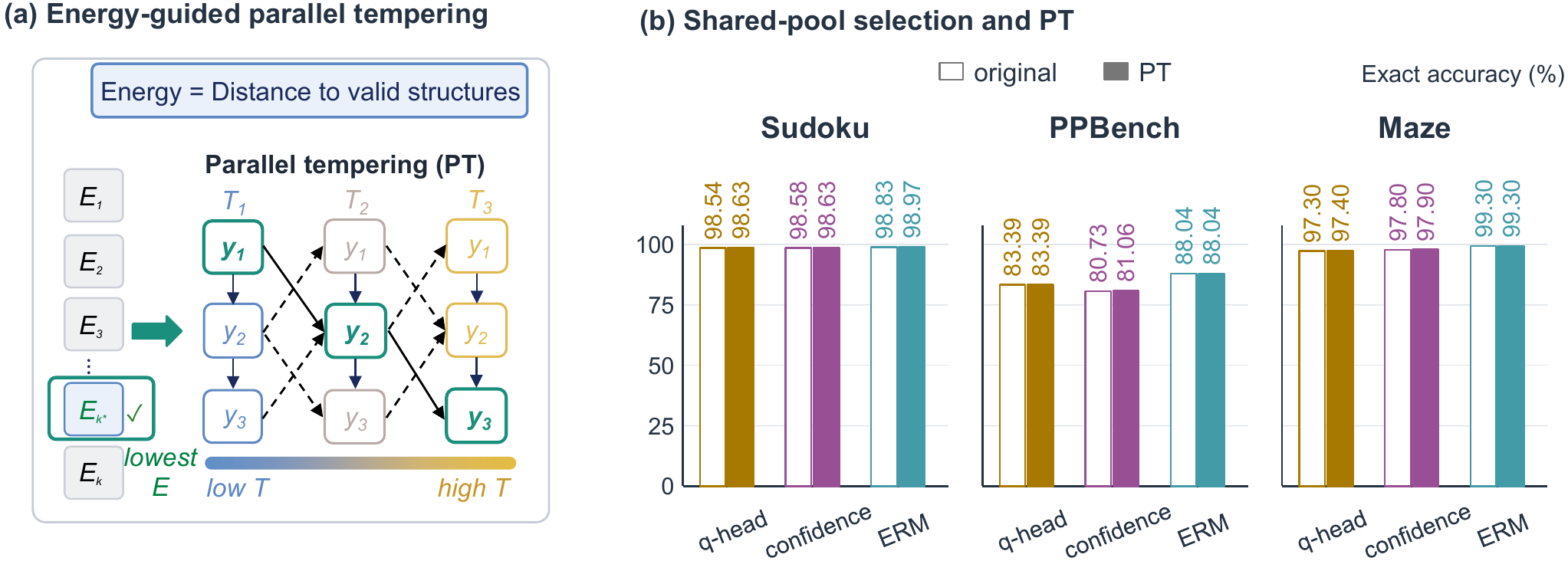}
  \caption{\textbf{Energy-guided parallel tempering and selection.} (a) The task energy guides exchanges between neighboring temperatures and selects the lowest-energy candidate. Vertical arrows show refinement; diagonal arrows show exchange proposals accepted by Equation~\ref{eq:pt_swap}. Teal states illustrate a replica moving across temperatures. (b) Hollow bars show shared-pool selection; solid bars show PT with the same scoring energy. All values are exact accuracy (\%).}
  \label{fig:pt_energy}
\end{figure}

\begin{table}[t]
\caption{\textbf{Parallel tempering with different energies.}
Original is the accuracy of the same selector on the shared EqR candidate pool, selected accuracy is the final answer chosen after PT sampling, and oracle accuracy is pass@$K$ within the PT-generated candidates.}
\label{tab:pt}
\centering
\small
\setlength{\tabcolsep}{3.5pt}
\begin{tabular}{llrrrrr}
\toprule
Task & PT energy & Original & Selected & Oracle & Swap rate & Count \\
\midrule
Sudoku & q-head & 98.54 & 98.63 & 98.68 & 0.7964 & 2020/2048 \\
Sudoku & Confidence & 98.58 & 98.63 & 98.63 & 0.9548 & 2020/2048 \\
Sudoku & ERM & 98.83 & \textbf{98.97} & 98.97 & 0.9010 & 2027/2048 \\
\midrule
PPBench & q-head & 83.39 & 83.39 & 87.38 & 0.8948 & 251/301 \\
PPBench & Confidence & 80.73 & 81.06 & 88.04 & 0.9939 & 244/301 \\
PPBench & ERM & 88.04 & \textbf{88.04} & 88.04 & 0.9378 & 265/301 \\
\midrule
Maze & q-head & 97.30 & 97.40 & 99.20 & 0.8669 & 974/1000 \\
Maze & Confidence & 97.80 & 97.90 & 99.30 & 0.9952 & 979/1000 \\
Maze & ERM & 99.30 & \textbf{99.30} & 99.30 & 0.9471 & 993/1000 \\
\bottomrule
\end{tabular}
\end{table}

\subsection{Sampling Extension: Parallel Tempering}

PT-ERM raises Sudoku accuracy from $98.83\%$ to $98.97\%$ ($2027/2048$), adding three solves, and preserves the PPBench and Maze results of $88.04\%$ and $99.30\%$ (Figure~\ref{fig:pt_energy}; Table~\ref{tab:pt}). Selected accuracy matches each PT pool's oracle. The main gains come from selection; the three-solve PT improvement is modest, and its robustness across random seeds remains untested.

\subsection{Improved Language Modeling}
\label{sec:language_results}

\begin{figure}[!ht]
\centering
\includegraphics[width=\linewidth]{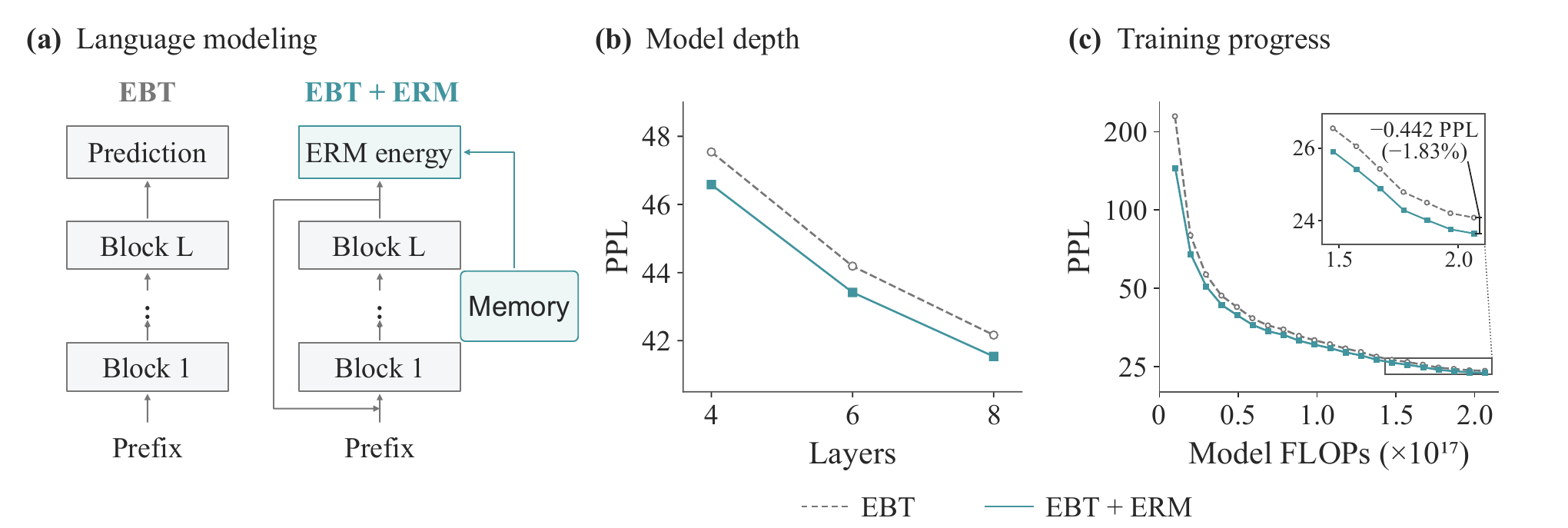}
\caption{\textbf{ERM for language modeling.} (a) EBT backbone and ERM inference with continuation memory. (b) Perplexity (PPL) of 4-, 6-, and 8-block models on the 33k-document proxy after 105,000 training steps. (c) Six-block PPL across 21 checkpoints on a fixed 256-document slice, against pretraining plus candidate-generation FLOPs (one EBT candidate; eight ERM candidates). The main PPL axis is logarithmic; the linear-scale inset shows steps 75,000--105,000 and the final-checkpoint PPL reduction. EBT denotes the first candidate; EBT + ERM uses Equation~\ref{eq:language_energy} with $K=8$.}
\label{fig:language_ebt_erm}
\end{figure}

Language modeling tests the same energy-guidance principle using Hopfield-type memories of observed continuations. Figure~\ref{fig:language_ebt_erm}(a) shows its integration with a pretrained EBT. We use an S2 EBT with six blocks and 44.86M parameters, trained for 105,000 steps, and evaluate $K=8$ shared trajectories on a frozen RedPajama-V2 proxy of 33,000 documents and 7,397,883 target tokens \citep{gladstone2026ebt,weber2024redpajama}. The continuation memory combines training-source suffix counts with earlier observations in the current document. We report token-weighted perplexity, $\exp(\sum_d\mathrm{NLL}_d/\sum_d n_d)$, where $n_d$ counts scored tokens; Appendix~\ref{app:language} gives the evaluation settings.

\begin{table}[ht]
\caption{\textbf{Language modeling with a fixed EBT.} All methods use the same $K=8$ candidates on the 33k proxy. Lower perplexity is better. ERM uses Equation~\ref{eq:language_energy}; the learned scalar energy selects one trajectory.}
\label{tab:language}
\centering\small
\begin{tabular}{lrr}
\toprule
Method & Perplexity & Reduction (\%) \\
\midrule
First candidate & 44.189 & --- \\
Candidate average & 44.147 & 0.094 \\
EBT scalar-energy selection & 44.180 & 0.020 \\
\textbf{ERM Hopfield energies} & \textbf{43.419} & \textbf{1.742} \\
\bottomrule
\end{tabular}
\end{table}

ERM reduces perplexity from $44.189$ to $43.419$ (Table~\ref{tab:language}), a $1.742\%$ improvement (paired-document bootstrap 95\% CI: $1.711$--$1.773\%$), and improves on candidate averaging by $1.649\%$. Selecting the trajectory with the lowest summed EBT scalar energy reduces perplexity by $0.020\%$. Hopfield energies also lower PPL from $47.541$ to $46.581$ for four blocks and from $42.166$ to $41.538$ for eight blocks (Figure~\ref{fig:language_ebt_erm}(b)). These comparisons show consistent gains across the three backbone depths with fixed pretrained parameters.

Across 21 checkpoints (steps 5,000--105,000), we hold the memory, hyperparameters, and 256-document slice fixed. Mean next-token energy decreases at every interval, from $4.977$ to $3.163$ nats; the corresponding PPL falls from $145.033$ to $23.641$ (Figure~\ref{fig:language_ebt_erm}(c)).

ERM adds no optimizer steps or trainable parameters and uses the same eight trajectories as candidate averaging. Its additional cost is memory construction, lookup, and normalization. The full 33k comparison takes $2.83$ hours on one RTX 4090; Appendix~\ref{app:language} gives the timing scope and development cost.

\subsection{Energy-Based Early Stopping}
\label{sec:energy_stopping}

The energy dynamics in Figure~\ref{fig:maze_depth_curve} suggest that task compatibility can guide recurrent depth as well as final selection. On Maze, we monitor the batch mean of the lowest candidate energy for each input and stop when its relative change stays below $0.01$ for two consecutive checks, starting at step four. With $K=128$ PT replicas and a maximum depth of $64$, all 63 batches stop by this criterion at depths $6$--$11$, averaging $7.60$ steps, with $996/1000=99.60\%$ exact accuracy (Table~\ref{tab:maze_stopping}). Appendix~\ref{app:maze_early_stop} gives the precise criterion.

Task energy supplies a batch-level stopping signal without additional ACT training~\citep{graves2016adaptive}. The runs use different pools, and reduced depth does not by itself establish lower wall-clock cost.

\begin{table}[ht]
\caption{\textbf{Maze accuracy with reduced recurrent depth.} All runs use $K=128$ and the same 1000-example test split. The early-stopping depth is averaged over 63 batches; the adaptive run uses PT and is not a shared-pool comparison with the fixed-depth runs.}
\label{tab:maze_stopping}
\centering\small
\begin{tabular}{lrrrr}
\toprule
Inference & Max. depth & Mean stop & Exact (\%) & Count \\
\midrule
ERM, fixed depth & 64 & --- & 99.30 & 993/1000 \\
ERM, fixed depth & 16 & --- & 99.90 & 999/1000 \\
ERM + PT, energy stopping & 64 & 7.60 & 99.60 & 996/1000 \\
\bottomrule
\end{tabular}
\end{table}

\section{Discussion}

ERM uses one principle across reasoning and language: Hopfield-type memories define Hopfield energies that assess candidate predictions. Reasoning uses valid patterns and global constraints; language uses observed continuations. This framework recovers candidate-pool oracle accuracy on reasoning tasks and improves language perplexity without further training. On Maze, the energies also guide stopping, reaching $99.60\%$ accuracy at a mean depth of $7.6$. More recurrence is therefore not always better.

Recurrent computation is also being explored in billion-parameter language models, including recurrent-depth transformers and Ouro's looped models \citep{geiping2025latent,zhu2025ouro}. Hopfield energies could help combine their predictions or guide stopping when energy tracks prediction quality. Our fixed-EBT experiment demonstrates the language formulation; transfer to larger recurrent language models remains untested.

\subsection{Limitations}

The framework is general, but its Hopfield-type memories and constraints depend on the task. Our reasoning comparisons do not isolate the local terms of Hopfield energies from a discrete rule checker or global penalties. Tasks with unknown rules require a way to infer suitable memories and constraints. Language modeling is evaluated on one corpus proxy.

Our language study isolates inference gains on a fixed iterative EBT, rather than comparing architectures with the nonrecurrent Transformer++ baseline \citep{gladstone2026ebt}. It therefore does not establish an advantage over Transformer++; that comparison requires matched training data and compute. Finally, stable energy does not certify correctness, and the stopping experiment does not directly compare against a trained ACT controller.

\section{Conclusion}

We presented ERM, a framework for energy guidance in recursive reasoning and language models. Hopfield-type memories define Hopfield energies for candidate selection, prediction, sampling, and depth control. ERM matches candidate-pool oracle accuracy on Sudoku, PPBench, and Maze, and reduces language-model perplexity by $1.74\%$ without further training. Maze stopping and the modest Sudoku gain from PT show how the same energies can guide computation. The results support energy guidance as a practical principle for test-time scaling, with gains depending on candidate quality and memory design.
\label{sec:main_end}

\subsection*{AI use statement}
Generative AI tools were used only to assist with writing and information search. They were not used for method development, experiment design, implementation, data processing, or result interpretation. The authors are responsible for verifying AI-assisted text and retrieved information and for the final content of this work.

\subsection*{Ethics statement}
This work evaluates ERM on structured puzzle benchmarks and a RedPajama-V2 language-modeling proxy, without human-subject experiments. Web-derived text may contain biased, harmful, or sensitive content; our evaluation does not establish privacy, fairness, or deployment-safety guarantees.

\subsection*{Reproducibility statement}
Section~3 defines the energies and inference procedure, and Section~4.1 describes the evaluation setup. Appendix~\ref{app:erm_pseudocode} provides inference pseudocode; Appendix~\ref{app:language} details the language protocol and memory construction; and Appendix~\ref{app:maze_early_stop} specifies the stopping criterion. Appendix~C reports parameter and compute accounting.

\bibliography{iclr2027_conference}
\bibliographystyle{iclr2027_conference}

\clearpage
\appendix

\section{Notation Table}

\small
\begin{longtable}{p{0.18\linewidth}p{0.74\linewidth}}
\caption{\textbf{Notation used in the main text.}}
\label{tab:notation}\\
\toprule
Symbol & Meaning \\
\midrule
\endfirsthead
\toprule
Symbol & Meaning \\
\midrule
\endhead
$x$ & Input puzzle or observed text prefix. \\
$q$ & Query state in the Modern Hopfield retrieval energy. \\
$\lVert q\rVert^2$ & Squared Euclidean norm of the Modern Hopfield query state. \\
$D$ & Recurrent depth; reasoning experiments use $D\leq64$. \\
$K$ & Number of candidate rollouts or sampled trajectories. \\
$\mathcal{Y}(x)$ & Candidate pool produced for input $x$. \\
$y$ & Generic candidate output variable used when defining a distribution. \\
$y_k$ & The $k$-th candidate structured output. \\
$k$ & Candidate index inside the pool $\mathcal{Y}(x)$. \\
$\hat{k}$, $\hat{y}$ & Selected candidate index and selected structured answer. \\
$N$ & Number of output positions or cells. \\
$E(y_k;x)$ & Candidate energy; shorthand includes $L_k$ when required. Lower is better. \\
$E_\rho$, $E_\eta$ & Energies of two PT replicas proposed for swapping. \\
$E_{\mathrm{MHN}}$ & Modern Hopfield retrieval energy in Equation~\ref{eq:mhn_energy}. \\
$E_{\mathrm{ERM}}$ & ERM scoring function for Hopfield energies in Equation~\ref{eq:erm_energy}. \\
$E_{\mathrm{ERM}\mid\mathrm{Sudoku}}$, $E_{\mathrm{ERM}\mid\mathrm{PPBench}}$, $E_{\mathrm{ERM}\mid\mathrm{Maze}}$ & Task-specific instantiations of $E_{\mathrm{ERM}}$. \\
$J$ & Number of task factors used by ERM. \\
$j$ & Factor index, such as a Sudoku row or a PPBench clue neighborhood. \\
$\mathcal{M}$ & Generic memory set used in the Modern Hopfield retrieval energy. \\
$\mathcal{M}_j(x)$ & Input-conditioned memory set for factor $j$. \\
$m$ & One stored memory pattern in either $\mathcal{M}$ or $\mathcal{M}_j(x)$. \\
$d_j(y_k,m)$ & Factor mismatch, using $L_k$ for Sudoku; for language, $d_j(v,m)=-\log p_m(v)$. \\
$\tau$ & Soft-retrieval temperature for the memory log-sum-exp. \\
$G(y_k,x)$ & Global rule-violation score for constraints not enumerated locally. \\
$\mu$ & Weight of the global rule-violation term. \\
$T$ & Sampling temperature in the tempered distribution. \\
$\pi_T(y\mid x)$ & Tempered distribution over candidate outputs at temperature $T$. \\
$T_1,\ldots,T_R$ & Parallel-tempering replica temperatures. \\
$T_\rho$, $T_\eta$ & Temperatures of two PT replicas proposed for swapping. \\
$R$ & Number of PT replicas. \\
$\rho$, $\eta$ & Replica indices used in the PT swap rule. \\
$a$ & Swap acceptance probability between two PT replicas. \\
$i$, $d$ & Output position and digit/token index. \\
$L_k$, $L_k[i,d]$ & Predictive log-probability tensor from rollout $k$ and its entry at position $i$, digit/token $d$. \\
$m_i$ & Digit assigned by Sudoku memory $m$ to cell $i$. \\
$\mathrm{sim}$ & Generic similarity function used in Modern Hopfield retrieval. \\
$\beta$ & Scale parameter for memory similarity or distance. \\
$t$, $x_{<t}$ & Token position and observed prefix in language prediction. \\
$v$, $u$, $\mathcal V$ & Candidate token, summation token, and vocabulary. \\
$p_k(v)$, $p_0(v)$ & Candidate token distribution and uniform candidate average. \\
$q_t(v)$, $g_t$, $r(v)$ & Continuation memory, memory gate, and interpolated distribution. \\
$w_{j,m}$, $p_m(v)$ & Memory weight and token distribution for memory $m$ in factor $j$. \\
$E_{\mathrm{LM}}$, $Z_t$ & Language energy and its exact normalization constant. \\
$n_d$, $\mathrm{NLL}_d$ & Scored token count and summed negative log-likelihood for document $d$. \\
\bottomrule
\end{longtable}

\normalsize

\subsection{ERM Inference Pseudocode}
\label{app:erm_pseudocode}

The procedure below describes independent rollouts followed by ERM selection; it does not include the optional PT exchanges.  EqR supplies the stochastic depth--breadth candidate pool, following the randomized-initialization and repeated-update structure in Algorithm 1 of \citet{huang2026eqr}.  ERM leaves these neural updates unchanged and selects candidates using Hopfield energies from Equation~\ref{eq:erm_energy}.

\begin{center}
\begin{minipage}{0.96\linewidth}
\small
\hrule
\vspace{2pt}
\textbf{Algorithm 1: ERM selection from independent rollouts}\\[-1pt]
\hrule
\vspace{3pt}
\begin{tabbing}
\quad\=\quad\=\quad\=\kill
\textbf{Input:} input $x$; EqR update $f_\theta$; depth $D$; candidates $K$;\\
\> task memories $\{\mathcal{M}_j(x)\}_{j=1}^J$; global score $G$;\\
\> task scoring parameters, including $\tau$, $\beta$, and $\mu$.\\
\textbf{Output:} selected structured answer $\hat{y}$.\\[2pt]
\textbf{for} $k=1,\ldots,K$ \textbf{do}\\
\> initialize the EqR carry for candidate $k$ independently.\\
\> \textbf{for} $t=1,\ldots,D$ \textbf{do}\\
\>\> update the carry with $f_\theta$ conditioned on $x$.\\
\> \textbf{end for}\\
\> decode $y_k$ and retain predictive log-probabilities $L_k$.\\
\textbf{end for}\\[2pt]
\textbf{for} $k=1,\ldots,K$ \textbf{do}\\
\> score $y_k$ using the task memories, global score, and parameters;\\
\> for Sudoku, use $L_k$ in Equation~\ref{eq:sudoku_energy}.\\
\textbf{end for}\\
$\hat{k}\leftarrow \arg\min_{k\in\{1,\ldots,K\}} E_{\mathrm{ERM}}(y_k;x)$; \quad $\hat{y}\leftarrow y_{\hat{k}}$.\\
\textbf{return} $\hat{y}$.
\end{tabbing}
\vspace{1pt}
\hrule
\end{minipage}
\end{center}

Here $x$ is the input instance, $f_\theta$ is the trained EqR recurrent update, $D$ is the number of recurrent steps, and $K$ is the number of independent rollouts.  The indices $k$ and $t$ denote the candidate and recurrent step, $y_k$ is the final candidate from rollout $k$, $\mathcal{M}_j(x)$ is the valid memory set for factor $j$ among $J$ task factors, $G(y_k,x)$ measures global rule violations, $E_{\mathrm{ERM}}$ is the ERM energy, and $\hat{k},\hat{y}$ are the selected index and answer.

\section{Additional Experimental Details}

\subsection{EqR+ET Sudoku Motivation}

\begin{figure}[ht]
  \centering
  \includegraphics[width=0.72\linewidth]{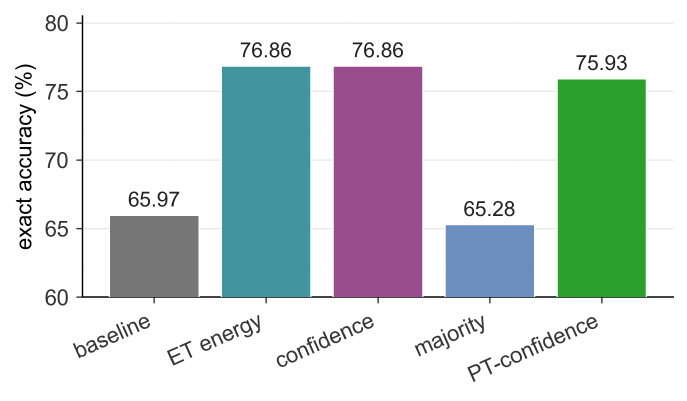}
  \caption{\textbf{Training with an Energy Transformer architecture.}
  This appendix experiment uses the EqR+ET Sudoku checkpoint with $D=64$, $K=16$, and PT temperatures from $0.001$ to $1.0$.  ET energy and confidence both improve over the one-rollout baseline, serving as motivation for the main inference-time ERM design, whereas PT-confidence falls slightly below direct confidence selection.}
  \label{fig:et_appendix}
\end{figure}

The EqR+ET Sudoku experiment compares selectors on a checkpoint trained with an Energy Transformer architecture. ET energy and confidence both solve $1574/2048=76.86\%$ of puzzles, compared with $1351/2048=65.97\%$ for one rollout. PT-confidence solves $1555/2048=75.93\%$, 19 fewer than direct confidence selection. These results motivate evaluating the scoring rule and sampling procedure separately. The main experiments use task memories with fixed pretrained generators and report selected and oracle accuracy for each candidate pool.

\begin{table}[ht]
\caption{\textbf{EqR+ET Sudoku appendix result.} Exact and token accuracy at $D=64$, $K=16$; PT-confidence uses confidence for sampling and final selection.}
\label{tab:et_appendix}
\centering
\small
\begin{tabular}{lrrr}
\toprule
Method & Exact & Token & Count \\
\midrule
baseline & 65.97 & 87.79 & 1351/2048 \\
ET energy & 76.86 & 91.20 & 1574/2048 \\
confidence & 76.86 & 91.25 & 1574/2048 \\
majority & 65.28 & 88.13 & 1337/2048 \\
PT-confidence & 75.93 & 90.99 & 1555/2048 \\
\bottomrule
\end{tabular}
\end{table}

\subsection{Language Protocol and Cost}
\label{app:language}

The S2 EBT has six independently parameterized blocks, width 384, six attention heads, and a 256-token context. It was pretrained with batch size 32 for 105,000 optimizer steps; the recorded duration is 62,960 seconds on an RTX PRO 6000. The language comparison uses one final checkpoint from a single training seed and a frozen proxy drawn from shards separate from training; this proxy was also used during development. Inference uses $K=8$, noise $0.5$, seed 33000, and batch size four. Confidence intervals quantify document-sampling uncertainty within this proxy using 3,000 paired bootstrap resamples of token-weighted NLL.

Continuation tables use 24,000 documents from 13 training sources, context lengths one to four, and up to 32 continuations per context. Document-local suffixes of lengths one to eight use only previously observed tokens, and the resulting memory keeps at most 192 continuations per position. The retained memory mass defines a gate capped at $0.95$ and scaled by $0.6$, selected on 256 calibration documents in a 1,280-document development run. The sparse correction includes an exact normalization over the vocabulary; no current or future target enters the energy. Its computation adds no gradient update or learned verifier.

The depth comparison evaluates 4-, 6-, and 8-block EBTs, each trained for 105,000 steps, on the full 33k proxy. The checkpoint sweep uses 256 documents (57,628 tokens), $K=8$, and fixed continuation tables and gate. In Figure~\ref{fig:language_ebt_erm}(c), pretraining costs $C(s)=s\times32\times256\times4.5\times(6\times2{,}457{,}600+38{,}597{,}376)$ FLOPs at optimizer step $s$, using batch size 32, context length 256, and 4.5 forward-equivalent training passes. The two inner terms count block and vocabulary-projection FLOPs. Two refinement steps plus a final forward pass cost seven forward equivalents per candidate on the doubled sequence, giving $I=57{,}628\times7\times(2\times6\times2{,}457{,}600+38{,}597{,}376)=2.747\times10^{13}$ FLOPs. We plot $C(s)+I$ for EBT and $C(s)+8I$ for ERM; first-candidate PPL comes from the shared $K=8$ evaluation, with its one-candidate cost assigned analytically. Model FLOPs exclude memory construction and lookup. Target tokens enter only the evaluation metric. The recorded development evaluation takes 390 seconds ($6.5$ minutes). Recorded evaluation times cover candidate generation and all comparison branches; separate memory construction and scoring times were not recorded.

\subsection{Task-Specific ERM Examples}

Sudoku ERM performs exact soft retrieval over the valid memories of each structural unit. The row, column, and box memory sets follow the task's known permutation rules, with clues fixed to the input digits. A log-permanent dynamic program sums the probability assigned to every valid assignment within each unit. Summing the resulting retrieval energies gives the Sudoku score in Equation~\ref{eq:sudoku_energy}.

PPBench ERM uses a local-global decomposition for its different puzzle families. Local factors cover enumerable rules such as clue counts, forbidden adjacencies, no-$2\times2$ patterns, and run-length memories. Global penalties account for connectivity and coverage. No separate confidence score is added to these task energies.

Maze ERM combines local path-degree memories with global path checks. The local factors score allowed patterns around terminal, open, and wall cells. The global term checks format, connectivity, degree, cycle, and length violations in the proposed path. These checks use the input maze and candidate assignment, without constructing a reference solution. Together, the terms evaluate the local and global constraints of a valid start-to-goal path.

\subsection{Maze Depth Reduction and ERM Energy Stopping}
\label{app:maze_early_stop}

The Maze energy curve in Figure~\ref{fig:maze_depth_curve} motivates evaluating smaller recurrent budgets. Table~\ref{tab:maze_d16_shared} reports $D=16$ and $K=128$ on the same 1000-example test split. ERM selects $999/1000=99.90\%$ exact solutions and matches this pool's oracle, while q-head and confidence select $994/1000=99.40\%$. The smaller depth therefore yields a pool with high solution coverage that ERM fully recovers.

\begin{table}[ht]
\caption{\textbf{Maze shared-candidate selection at $D=16$, $K=128$.}  Exact and token accuracy are reported on the official 1000-example Maze-Unique test split.}
\label{tab:maze_d16_shared}
\centering
\small
\begin{tabular}{lrrr}
\toprule
Method & Exact & Token & Count \\
\midrule
Baseline & 81.40 & 98.20 & 814/1000 \\
q-head & 99.40 & 100.00 & 994/1000 \\
Confidence & 99.40 & 99.99 & 994/1000 \\
ERM & \textbf{99.90} & 99.99 & 999/1000 \\
Oracle & \textbf{99.90} & 99.99 & 999/1000 \\
\bottomrule
\end{tabular}
\end{table}

For the adaptive run in Table~\ref{tab:maze_stopping}, PT maintains $K=128$ replicas per input. At step $t$, the stopping statistic is the batch mean of the minimum candidate energy,
\[
s_t=\frac{1}{B}\sum_{b=1}^{B}\min_k E(y_{b,k}^{(t)};x_b),\qquad
\delta_t=\frac{|s_t-s_{t-1}|}{\max(|s_{t-1}|,10^{-6})}.
\]
Checks begin at $t=4$; the whole batch stops after two consecutive checks with $\delta_t<0.01$, or at the maximum depth $64$. The test set is processed in 63 batches of at most 16 examples. All batches stop by energy stability, with an unweighted mean depth of $7.603$ (standard deviation $1.148$, range $6$--$11$). Exact accuracy is $99.60\%$ and token accuracy is $99.95\%$. The checkpoint and its existing halting head are retained, but no batch terminates through the model-halted branch. No additional halting controller is trained. Intermediate decoding and energy checks add computation, so this experiment establishes a reduction in recurrent steps, not measured wall-clock savings; it also does not isolate PT from stopping.

\section{Parameter and Compute Accounting}

We use a simple rollout-based compute count in Table~\ref{tab:param_compute}.  Here $D$ is the recurrent depth per trajectory, $K$ is the number of parallel candidates or restarts, and model evaluations is $D\times K$.  Equivalent layers multiply model evaluations by the per-step equivalent-layer count used in TRM/EqR: 42 for the Sudoku/PPBench-style backbones and 15 for the small Maze backbone.  ERM keeps the same neural rollout budget as the backbone and adds no learned neural parameters. These counts exclude memory retrieval, global-constraint evaluation, and repeated scoring for PT or stopping; they do not measure end-to-end latency. The last column records the final selector.

\begin{table}[ht]
\caption{\textbf{Parameter and compute accounting.}  Model evaluations count recurrent outer-step evaluations; equivalent layers follow the TRM/EqR accounting.}
\label{tab:param_compute}
\centering
\scriptsize
\setlength{\tabcolsep}{4pt}
\begin{tabular}{@{}lrrrrrl@{}}
\toprule
Model/task & Params (M) & $D$ & $K$ & Model evals. & Eq. layers & Extra selector \\
\midrule
TRM Sudoku & 5.00 & 16 & 1 & 16 & 672 & none \\
TRM attention & 7.00 & 16 & 1 & 16 & 672 & none \\
PTRM Sudoku & 5.00 & 64 & 100 & 6,400 & 268,800 & q-head \\
PTRM Maze & 7.00 & 16 & 100 & 1,600 & 67,200 & q-head \\
PTRM PPBench & 7.00 & 48 & 100 & 4,800 & 201,600 & q-head \\
EqR Sudoku & 5.03 & 64 & 128 & 8,192 & 344,064 & majority vote \\
EqR Maze & 2.64 & 64 & 128 & 8,192 & 122,880 & majority vote \\
ERM Sudoku & \textbf{5.03} & 64 & 128 & 8,192 & 344,064 & \textbf{Hopfield energies} \\
ERM PPBench & \textbf{7.12} & 64 & 128 & 8,192 & 344,064 & \textbf{Hopfield energies} \\
ERM Maze & \textbf{2.64} & 64 & 128 & 8,192 & 122,880 & \textbf{Hopfield energies} \\
\bottomrule
\end{tabular}
\end{table}
\section{Figure Source Data}

The plotting scripts export the numerical source data used in the figures as CSV files.  The Maze depth-curve figure uses per-depth selector energies, selected exact accuracies, and oracle diagnostics for q-head, confidence, and ERM on the official Maze-Unique candidate pool.  The PT figure uses completed Sudoku, PPBench, and Maze PT results, the shared-selector source table records the exact accuracies used in Table~\ref{tab:shared}, and the EqR+ET appendix figure records the five Sudoku-Lite exact accuracies shown in Figure~\ref{fig:et_appendix}.  These files are included to make figure values auditable against the corresponding experiment logs and result JSON files.

\end{document}